\documentclass{article}

\usepackage[final]{nips_2017}

\usepackage[utf8]{inputenc} 
\usepackage[T1]{fontenc}    
\usepackage{hyperref}       
\usepackage{url}            
\usepackage{booktabs}       
\usepackage{amsfonts}       
\usepackage{nicefrac}       
\usepackage{microtype}      
\usepackage{graphicx}
\graphicspath{{figures/}}
\usepackage{enumitem}
\usepackage{tabulary}
\usepackage{multirow}
\newcommand{\comment}[1]{}

\title{Learning and Visualizing Localized Geometric Features Using 3D-CNN: An Application to Manufacturability Analysis of Drilled Holes}

\author{
  Sambit Ghadai, Aditya Balu, Adarsh Krishnamurthy, Soumik Sarkar\\
  Department of Mechanical Engineering\\
  Iowa State University\\
  Ames, IA 50010 \\
  \texttt{sambitg|baditya|adarsh|soumiks@iastate.edu}
  \comment{\thanks{Use footnote for providing further
  		information about author (webpage, alternative
  		address)---\emph{not} for acknowledging funding agencies.} \\}
   \\
}

\begin{document}

\maketitle

\begin{abstract}
3D Convolutional Neural Networks (3D-CNN) have been used for object recognition based on the voxelized shape of an object. However, interpreting the decision making process of these 3D-CNNs is still an infeasible task. In this paper, we present a unique 3D-CNN based Gradient-weighted Class Activation Mapping method (3D-GradCAM) for visual explanations of the distinct local geometric features of interest within an object. To enable efficient learning of 3D geometries, we augment the voxel data with surface normals of the object boundary. We then train a 3D-CNN with this augmented data and identify the local features critical for decision-making using 3D GradCAM. An application of this feature identification framework is to recognize \emph{difficult-to-manufacture} drilled hole features in a complex CAD geometry. The framework can be extended to identify \emph{difficult-to-manufacture} features at multiple spatial scales leading to a real-time design for manufacturability decision support system.
\end{abstract}

\section{Introduction}
Deep learning (DL) algorithms, more specifically 3D-Convolutional Neural Networks (3D-CNN), hierarchically learn multiple levels of abstractions of the data. They have been extensively used in computer vision~\citep{SVRG14,lee2009convolutional,lore2015llnet,HY08}, specifically for object recognition. However, learning local features in a geometry is different from object recognition where the object is classified based on a collection of features. In this work, we make use of a semi-supervised methodology to learn localized geometric features of interest within the object based on the cost function of the overall object classification problem. For this purpose, we train a 3D-CNN to learn the key features of the object and also learn the variation in the features that can classify the object based on a given cost function. We make use of a voxelized 3D representation of the object augmented with surface normal information to identify these localized features.

One applications of the aforementioned methodology, which we explore in this paper, is to identify \textit{difficult-to-manufacture} features in a CAD geometry and ultimately, classify its manufacturability. There are different handcrafted design for manufacturability (DFM) rules that ensure manufacturability of a design. For this purpose, the hierarchical architecture of DL can be used to learn increasingly complex features by capturing \emph{localized geometric features} and \emph{feature-of-features}. Thus, a deep-learning-based design for manufacturing (DLDFM) tool can be used to learn the various DFM rules from different examples of manufacturable and non-manufacturable components without explicit handcrafting. 

A primary concern while examining the manufacturability of CAD geometries using a DL based approach is the black-box nature of such deep networks. Interpretation of the decision making process in the form of visual explanations is essential for extracting the local features in an object that effects its non-manufacturability. Visual detection of local features further enables re-designing of the component or object to abide the various DFM rules. Recent major work on interpreting DL output by~\citet{selvaraju2016grad} makes use of a 2D gradient-weighted class activation map for producing visual explanations of the CNN's decision making processes in object recognition in images. In this work, we extend the GradCAM to 3D objects for interpretation of 3D-CNN's outputs and visualizing the regions that give rise to non-manufacturability conditions in the objects.

In this paper, we present a 3D convolutional neural network (3D-CNN) based framework that will learn and identify localized geometric features from an expert database in a semi-supervised manner. Further, we present a visual explanation technique using GradCAM to interpret the decision making process in the context of manufacturability with different CAD models classified as \emph{manufacturable} and \emph{non-manufacturable}.

\begin{figure*}[!t]
	\centering
	\includegraphics[width=0.85\linewidth, clip, trim={0.0in 0.5in 0.0in 0in}]{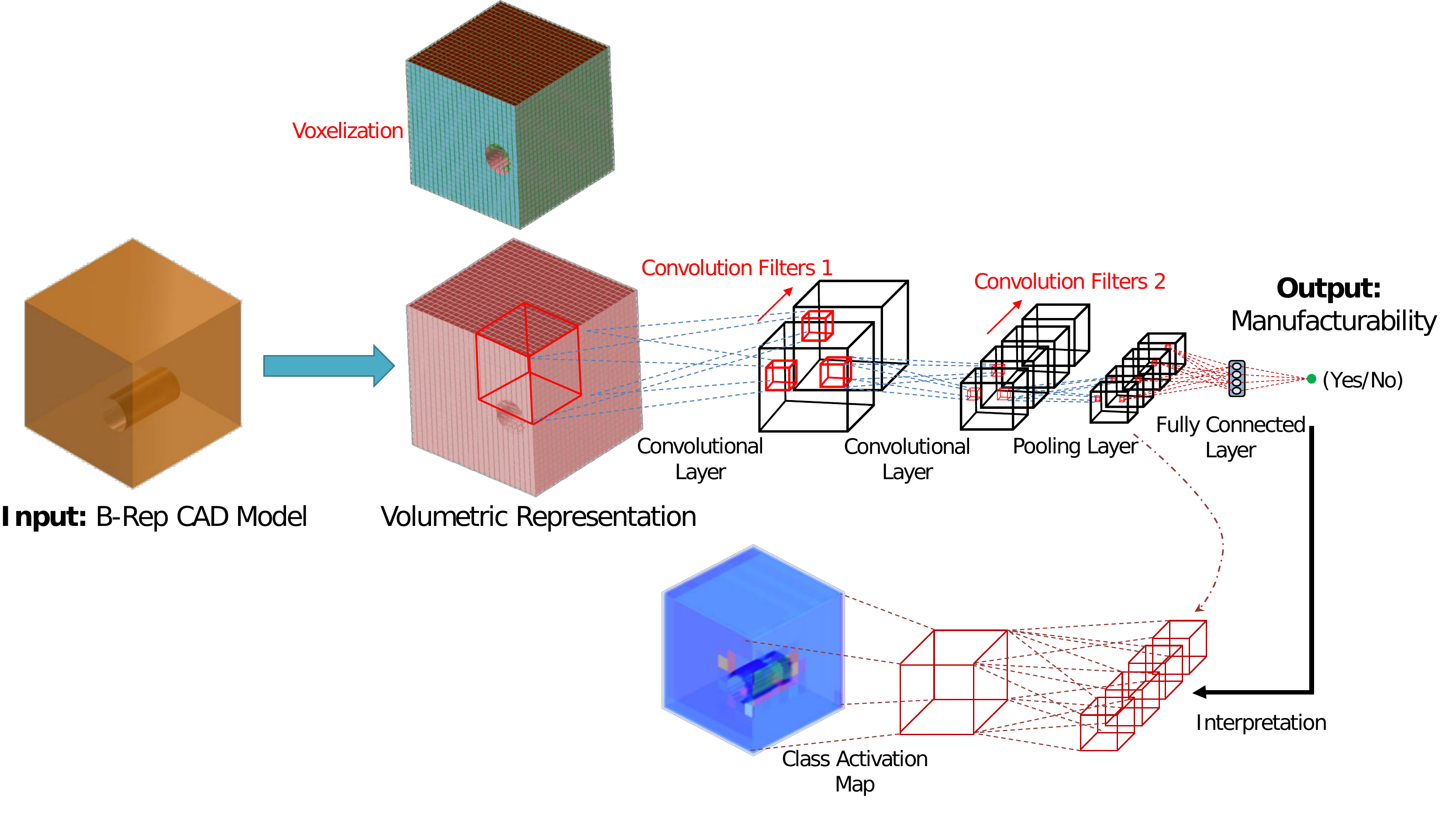}
	\caption{Framework for deep-learning based design for manufacturability. The CAD model is converted to a voxel representation and is input to a 3D-CNN for manufacturability classification. The 3D-CNN output is analyzed using 3D-gradCAM to provide manufacturability feedback.}
	\label{Fig:3DCNN}
\end{figure*}

\section{Volumetric Representations for Learning Geometric Features}
\label{Sec:VolumeRep}

Traditional CAD systems use boundary representations (B-Reps) to define and represent the CAD model~\citet{Krishnamurthy-2009-CAD}. In B-Reps, the geometry is defined using a set of faces that form the boundary of the solid object. B-Reps are ideally suited for displaying the CAD model by first tessellating the surfaces into triangles and using the GPU to render them. In our framework, we convert the B-Rep CAD model to a volumetric occupancy grid of voxels using a rendering-based approach. For representing a 3D object with more information regarding its geometry, we augment the boundary voxels with the surface normals of the B-Rep geometry. Finally,the $x, y,$ \& $z$ components of the surface normals are then embedded in the voxelization along with the occupancy grid to be used as \textit{four}-channel input to the 3D-CNN. 

\section{3D-CNN for Learning Localized Geometric Features}
\label{Sec:3DCNN}
The input to the 3D-CNN is a zero-padded voxelized CAD model. The convolution layer is activated with RELU function and is followed by batch normalization layer and a max. pooling layer. The same sequence of convolution, batch normalization and max. pooling is used again. A fully connected layer is used before the final output layer (manufacturability) with sigmoid activation. The model parameters $\theta$, comprised of weights $\textbf{W}$, and biases, $\textbf{b}$ are optimized by error back-propagation with binary cross-entropy loss function~\citep{hinton2006reducing} using the ADADELTA optimizer~\citep{zeiler2012adadelta}.

Several different CAD models were generated based on the DFM rules for drilling using ACIS~\citep{ACIS10}, a commercial CAD modeling kernel, for training the network. DFM rules for drilled holes are based on inter-related local geometric features such as the depth-to-diameter ratio~\citep{BoothroydDewhurst,bralla1999design} and the presence of thin walls surrounding the hole. These rules were used to classify the models as manufacturable or non-manufacturable.

\section{Interpretation of 3D-CNN Output}
\label{Sec:GradCAM}
The trained 3D-CNN network can be used to classify the manufacturability of any new geometry and can be treated as a black box. In a practical scenario, interpretability and explainability of the output provided by the 3D-CNN is essential. In this paper, we attempt to visualize the input features that lead to a particular output of the 3D-CNN and if possible, modify it. A similar approach was used in object recognition in images by using class activation maps to obtain class specific feature maps by~\citet{selvaraju2016grad}. The class specific feature maps could be obtained by taking a class discriminative gradient of the prediction with respect to the feature map for the class activation. In this paper, we present the first application of 3D gradient weighted class activation map (3D-GradCAM) for recognizing local features in a 3D object.

In order to get the feature localization map using 3D-GradCAM, we need to compute the spatial importance of each feature map $A_l$ in the last convolutional layer of the 3D-CNN, for a particular class, $c$ ($c$ can either be manufacturability or non-manufacturability, for the sake of generality) in the classification problem. This spatial importance for each feature map can be interpreted as weights for each feature map; it can be computed as the global average pooling of the gradients back from the specific class of interest as shown in Eqn.~\ref{Eqn:GlobalAvgPooling}.

The cumulative spatial importance activations that contribute to the class discriminative localization map, $L_{3DGradCAM}$, is computed using
\begin{equation}
\label{Eqn:GradCAM}
L_{3DGradCAM} = ReLU\left(\sum_l{\alpha_{l} \times A^l}\right),
\end{equation}
\noindent where $\alpha_{l}$ are the weights computed using
\begin{equation}
\label{Eqn:GlobalAvgPooling}
\alpha_{l} = \frac{1}{Z} \times \sum_i{ \sum_j{ \sum_k{ \frac{\partial y^c}{\partial A_{ijk}^l}}}}.
\end{equation}
\noindent We can compute the activations obtained for the input using $L_{3DGradCAM}$ to analyze the source of output. The heat map of $\left(L_{3DGradCAM}\right)$ is resampled using linear interpolation to match the input size, and then overlaid in 3D with the input to be able to spatially identify the source of non-manufacturability. This composite data is finally rendered using a volume renderer.

We make use of a GPU-based ray-marching approach to render this volume data. The rendering is parallelized on the GPU with each ray corresponding to the screen pixel being cast independently. The intersection of the ray with the bounding-cube of the voxel data is computed, and then the 3D volumetric data is sampled at periodic intervals. The sum of all the sampled values along the ray is then computed. This value is converted to RGB using a suitable color scale and rendered on the screen. Table~\ref{Tab:GradCAM} shows different volumetric renderings of the composite 3D-GradCAM data.

\section{Results and Discussion}
\label{Sec:Results}

\begin{table}[!t]
	\setlength\extrarowheight{2pt}
	\caption{Quantitative performance assessment of the DLDFM on test data sets.}\vspace{0.1in}
	\label{Tab:conf_test}
	\centering
	\small
	\newcommand\T{\rule{0pt}{2.7ex}}
	\newcommand\B{\rule[-1.3ex]{0pt}{0pt}}
	\newcommand{\tabincell}[2]{\begin{tabular}{@{}#1@{}}#2\end{tabular}}
	\tymin=.1in
	\tymax=2.5in 
	\begin{tabular}{lcccccc}
		\hline
		Test Data Type&\tabincell{c}{Model Description\B}&\tabincell{l}{True \T\\ Positive\B}&\tabincell{l}{True \T\\ Negative\B}&\tabincell{l}{False \T\\ Positive\B}&\tabincell{l}{False \T\\ Negative\B}&\tabincell{l}{Accuracy\B}\\   
		\hline                
		\multirow{2}{*}{\tabincell{l}{\\675 models\\408 Manufacturable}}& \tabincell{c}{In-outs\T\B} & 391& 90 & 17 & 176& 0.7136\\
		\cline{2-7}
		&\tabincell{c}{In-outs + \\Surface Normals \T\B}& 334& 201 & 74 & 65& \textbf{0.7938}\\
		\hline
		
	\end{tabular}
	
\end{table}

The generated CAD geometries are classified to be manufacturable or non-manufacturable based on the DFM rules for drilled holes. The 3D-CNN (explained in Section~\ref{Sec:3DCNN}) is trained on the generated data with fine-tuned hyper-parameters to have the least validation loss. The architecture of the DLDFM network with voxelized information is composed of three convolution layers with filter sizes of 8, 4, and 2 respectively. Likewise, the DLDFM networks using surface normal along with voxelized representation, comprises of three convolution layers with filter sizes of 6, 3, and 2 respectively. In succession to the first and last Convolution layers, we use MaxPooling layers of subsampling size 2. A batch size of 64 is selected while training the DLDFM networks. The training was performed using Keras~\citep{chollet2015keras} with a TensorFlow~\citep{tensorflow2015-whitepaper} backend in Python environment. The training is performed until the validation loss remains constant for at least 10 consecutive epochs.

After successful training, the DLDFM network was tested on a test set to benchmark its performance. Accuracy of DLDFM network on the test set using the two data representations is shown in Table~\ref{Tab:conf_test}. The test-set has completely different geometries compared to the training set. Thus, it can be seen that the DLDFM is learning the \textit{localized geometric features}.


\begin{table}[!b]
	\caption{Illustrative examples of manufacturability prediction and interpretation using the DLDFM framework.}\vspace{0.1in}
	\label{Tab:GradCAM}
	\centering
	\small
	\newcommand\T{\rule{0pt}{2.7ex}}
	\newcommand\B{\rule[-1.3ex]{0pt}{0pt}}
	\newcommand{\tabincell}[2]{\begin{tabular}{@{}#1@{}}#2\end{tabular}}
	\tymin=.1in
	\tymax=2.5in 
	\begin{tabular}{lcccccc}
		\hline
			CAD Models&
			\tabincell{c}{\\\includegraphics[width=0.17\linewidth,trim={16mm 6mm 16mm 6mm},clip]{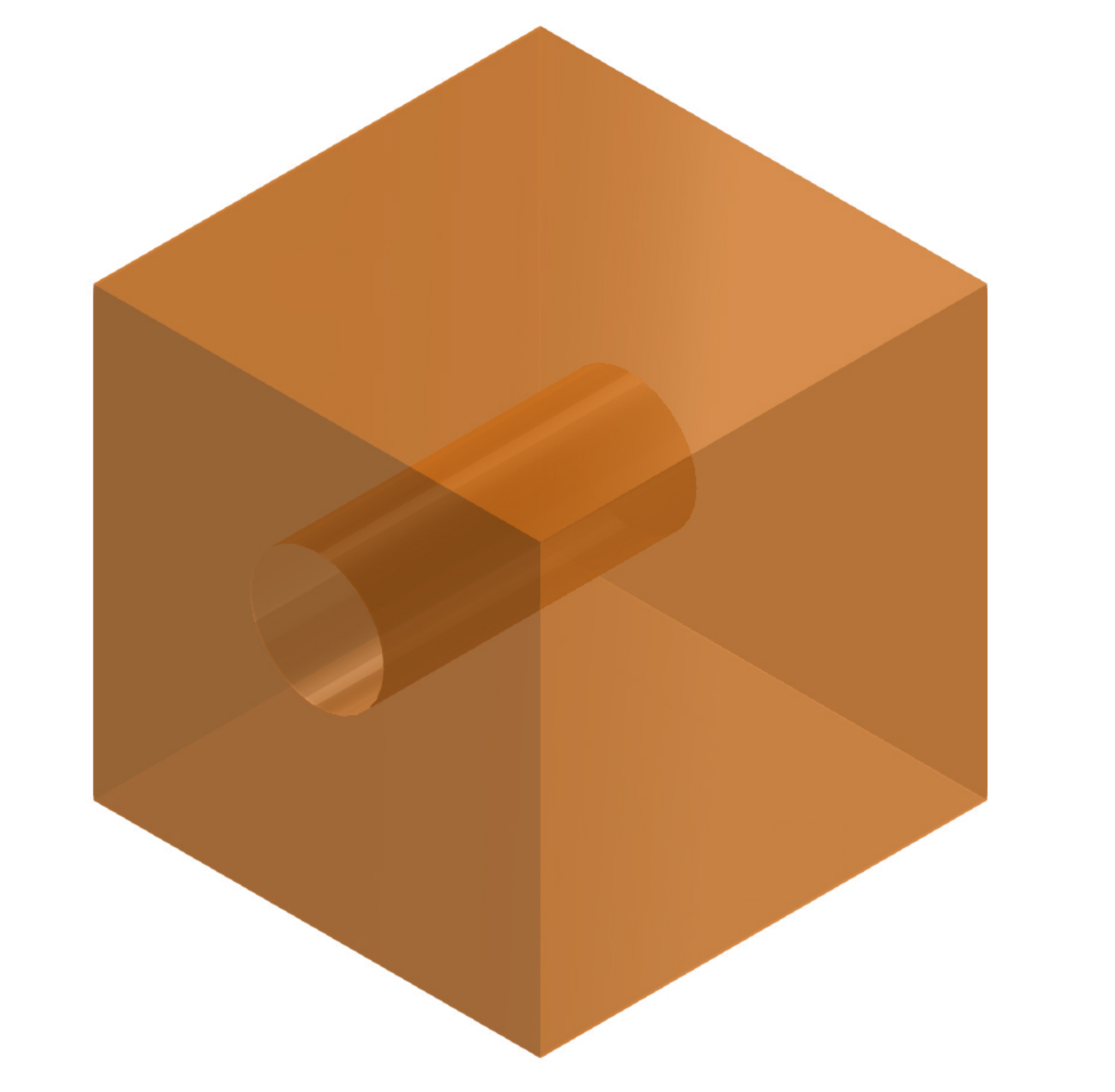}\\Manufacturable}&
			\tabincell{c}{\\\includegraphics[width=0.16\linewidth,trim={16mm 6mm 16mm 3mm},clip]{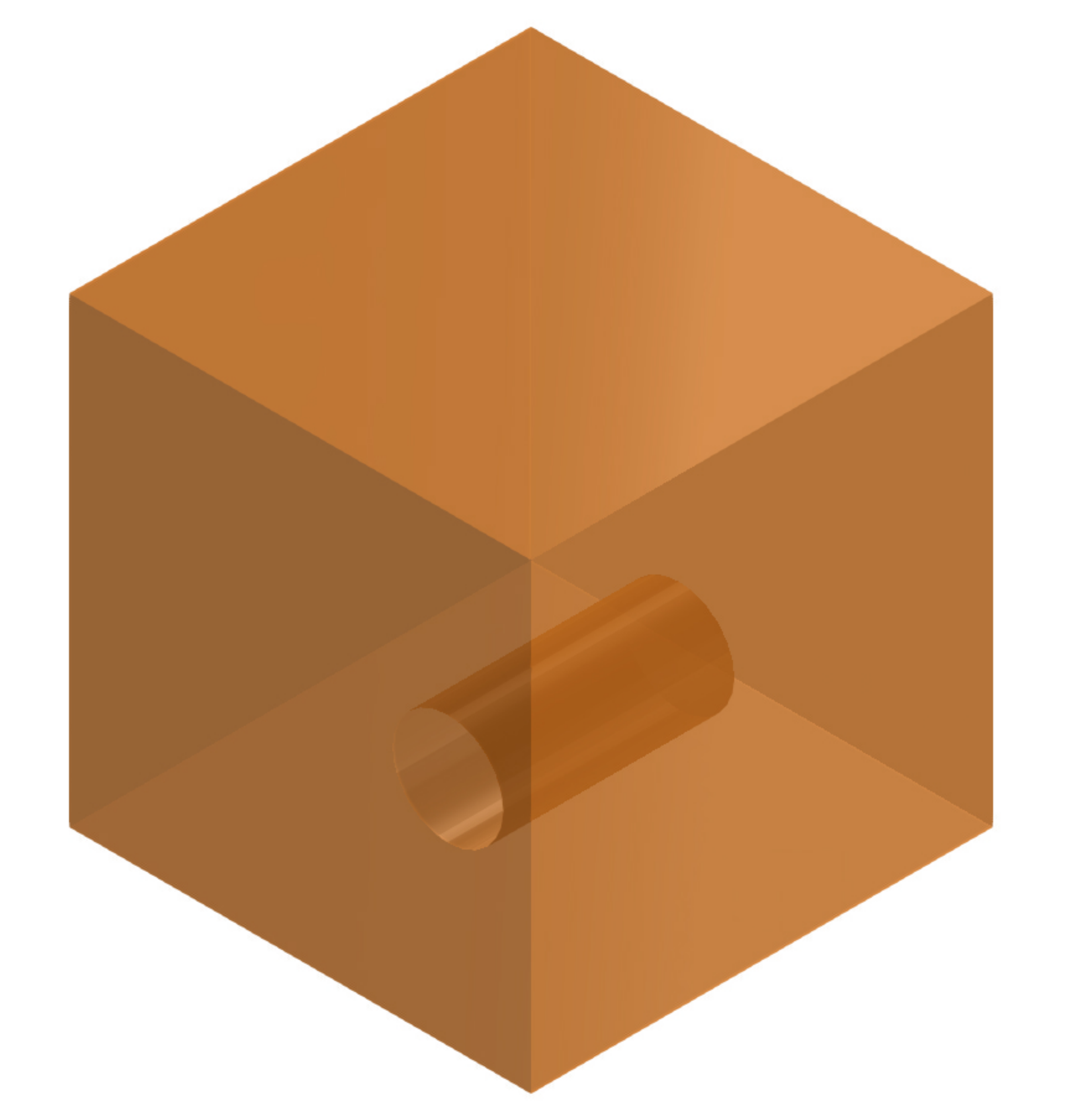}\\Non-Manufacturable}&
			\tabincell{c}{\\\includegraphics[width=0.16\linewidth,trim={16mm 3mm 16mm 3mm},clip]{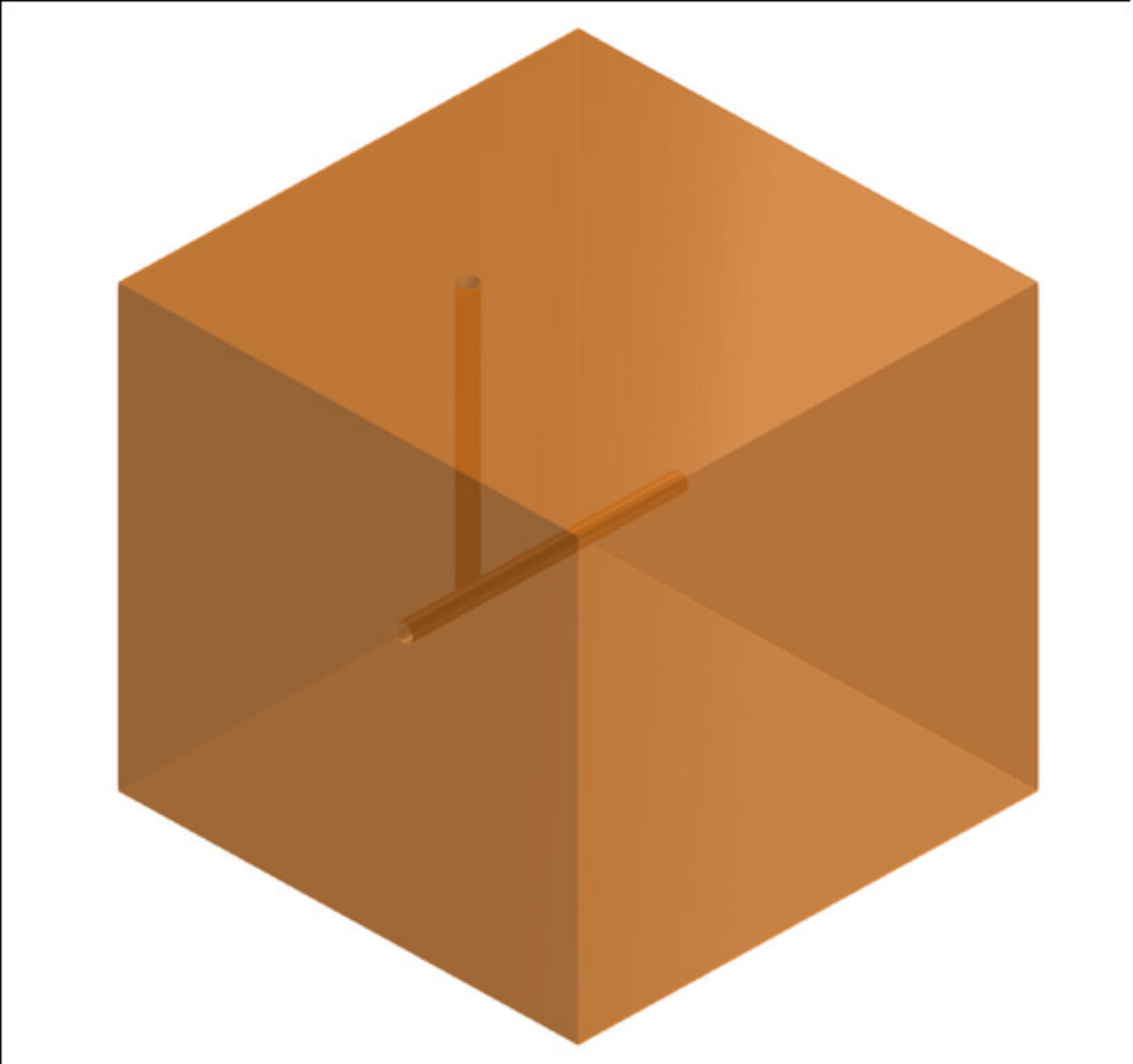}\\Non-Manufacturable}&
			\tabincell{l}{\\\includegraphics[width=0.17\linewidth,trim={16mm 8mm 16mm 4mm},clip]{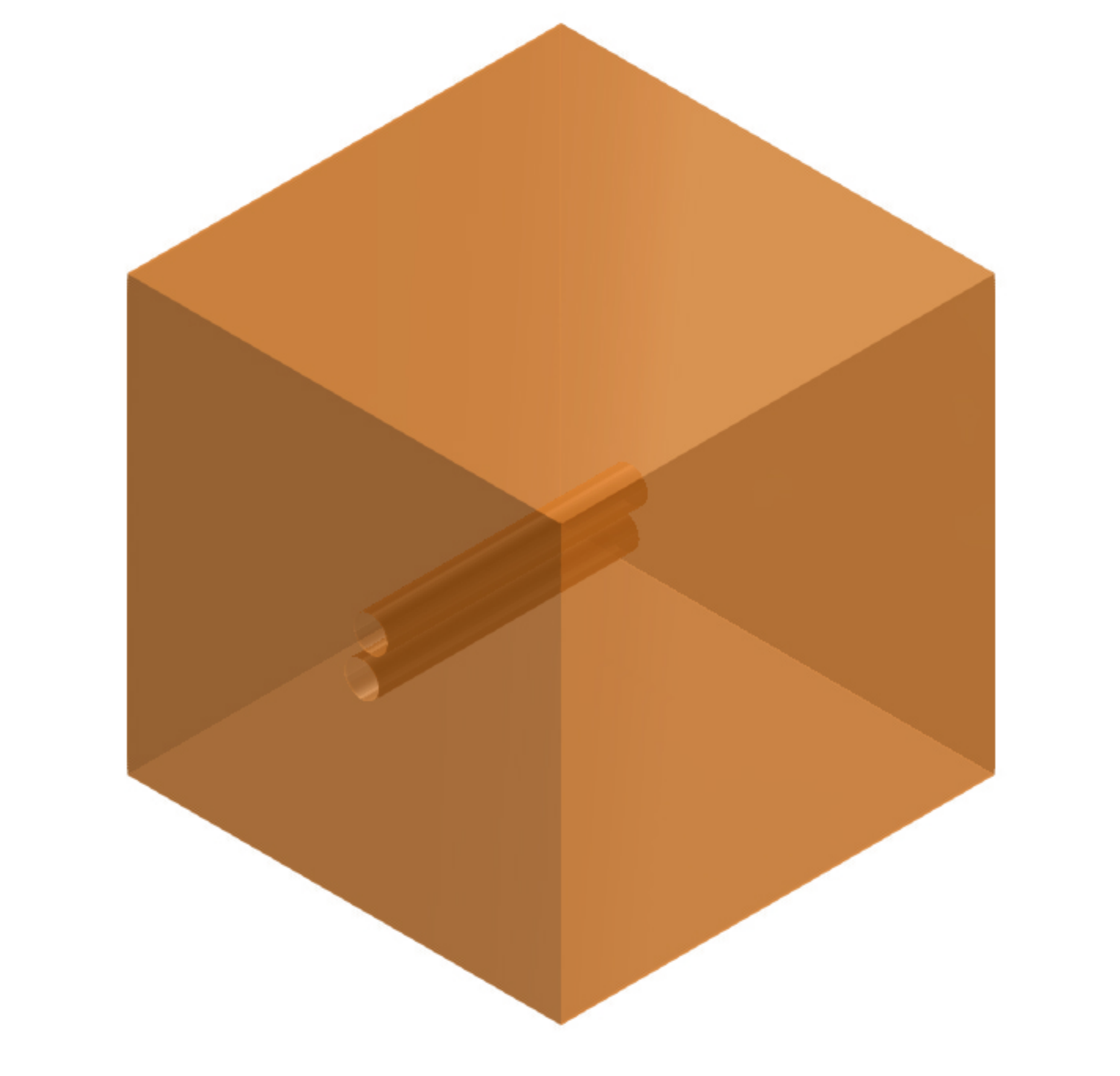}\\Non-Manufacturable}\\
			\hline                
			\multirow{2}{*}{\tabincell{l}{\\3D-GradCam\\Visualizations}}& \tabincell{c}{\\\includegraphics[width=0.18\linewidth,trim={16mm 6mm 16mm 6mm},clip]{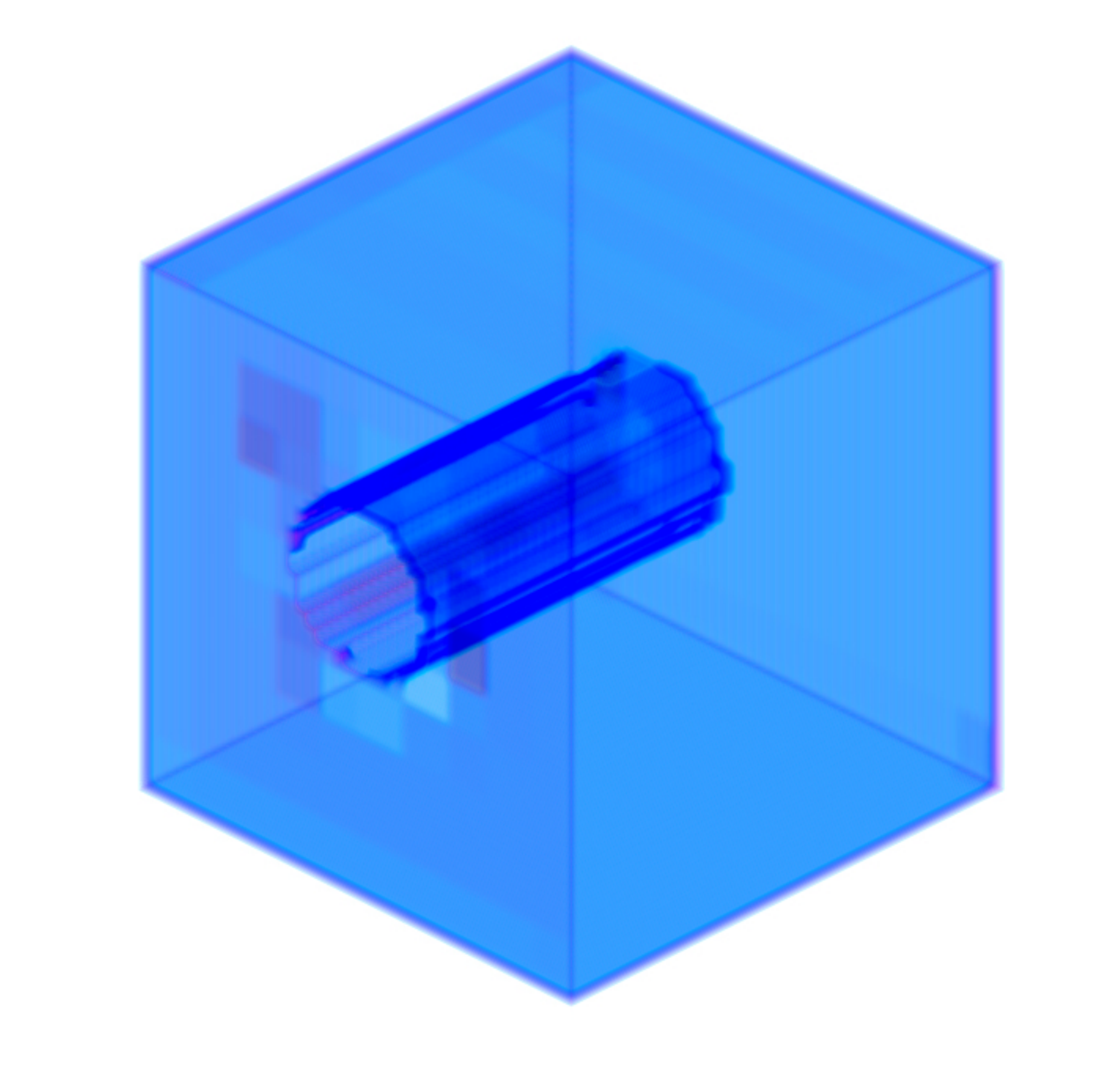}}&
			\tabincell{c}{\\\includegraphics[width=0.18\linewidth,trim={16mm 6mm 16mm 6mm},clip]{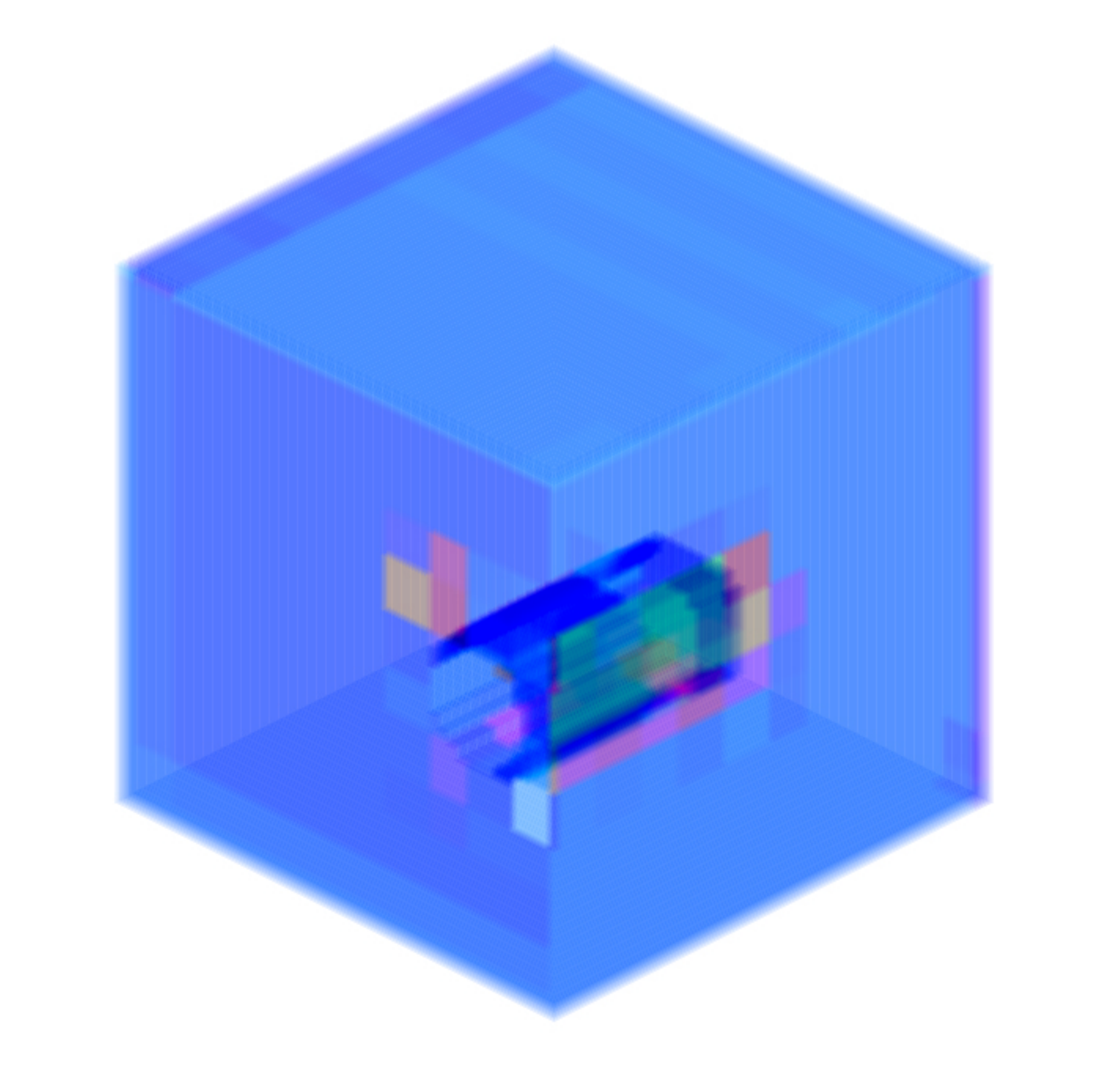}}&
			\tabincell{c}{\\\includegraphics[width=0.16\linewidth,trim={16mm 6mm 16mm 6mm},clip]{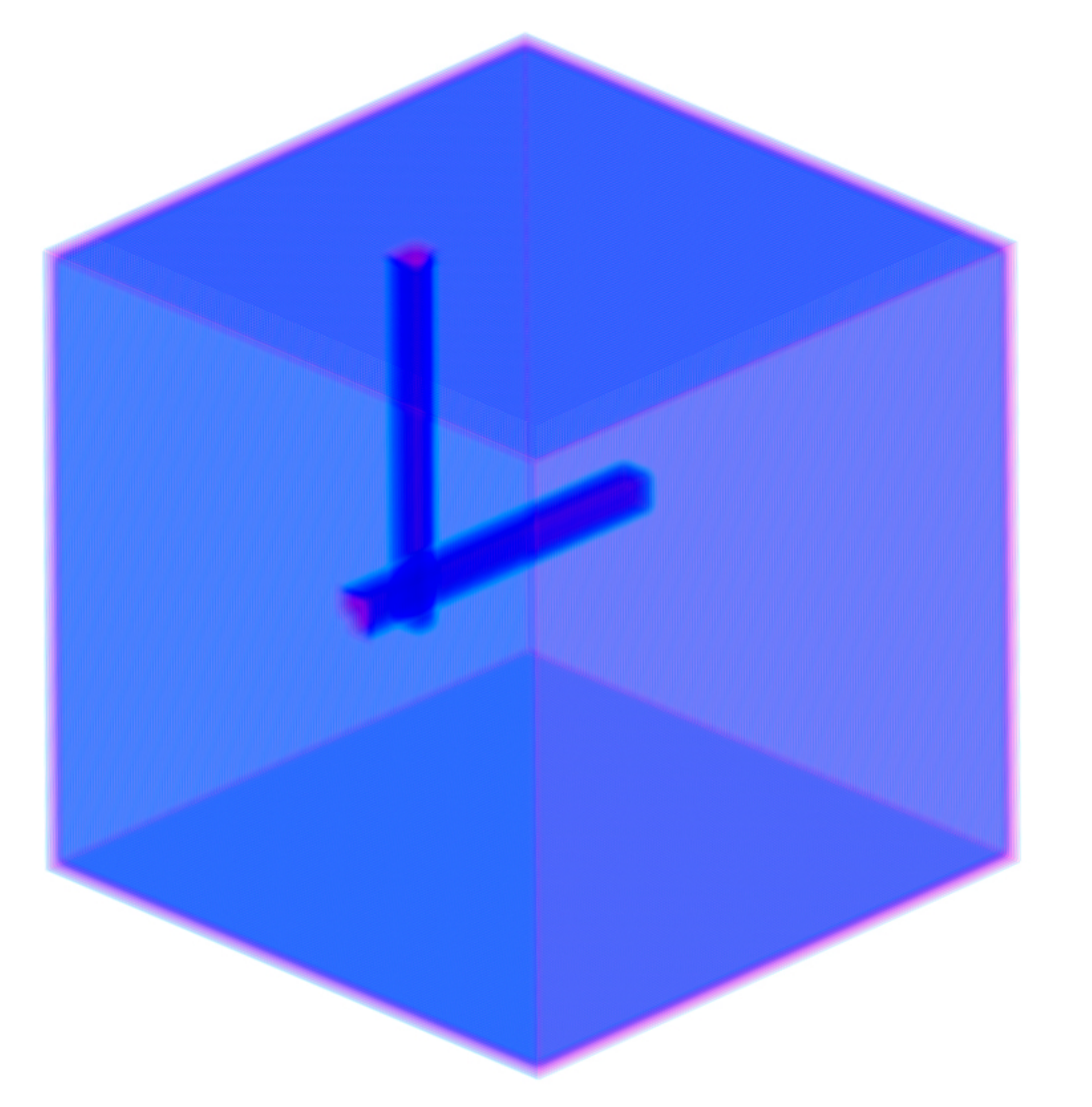}}&
			\tabincell{c}{\\\includegraphics[width=0.16\linewidth,trim={16mm 6mm 16mm 6mm},clip]{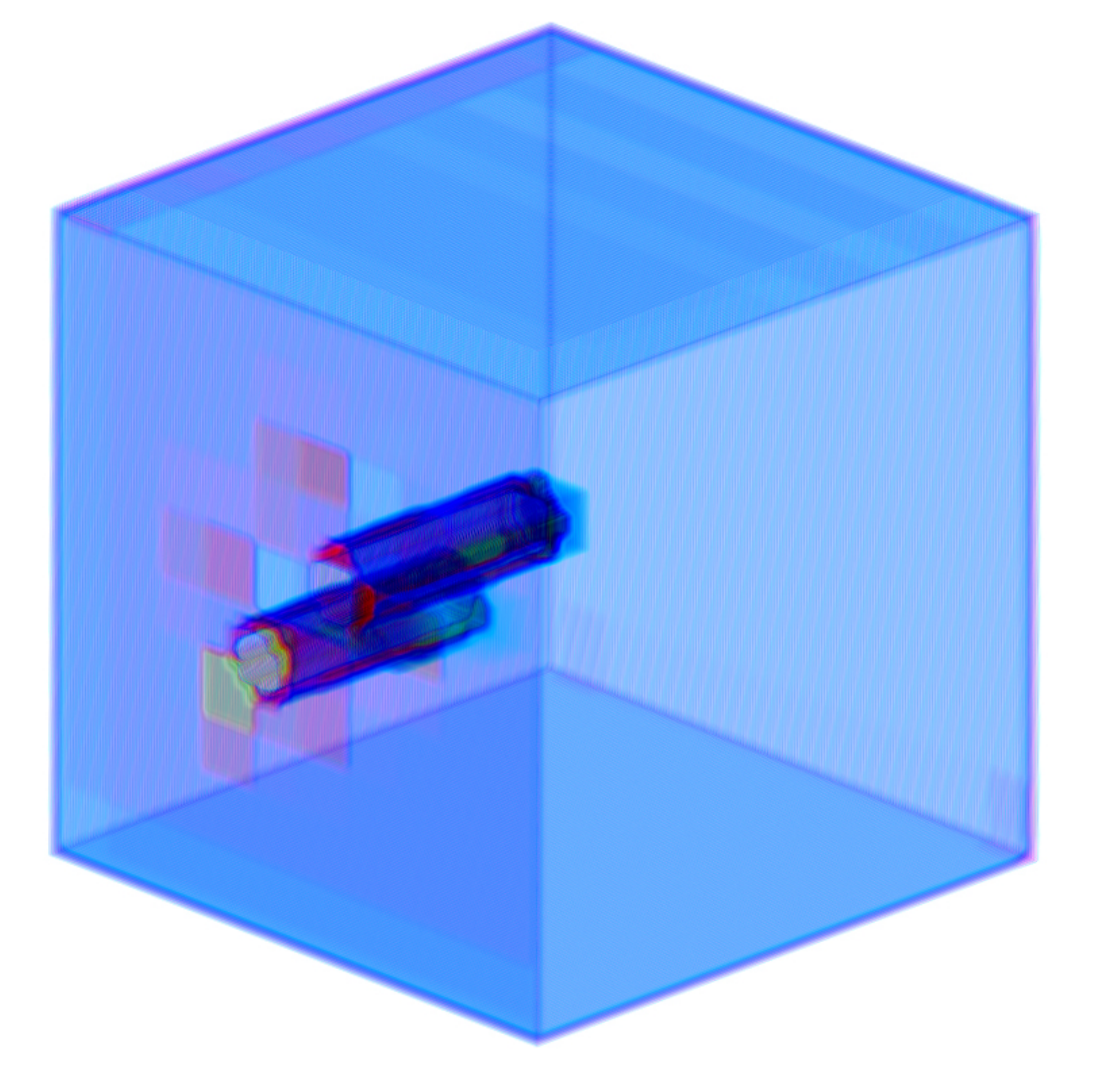}} \\
			\multicolumn{5}{c}{\includegraphics[width=0.39\linewidth,trim={0mm 0mm 0mm 2mm},clip]{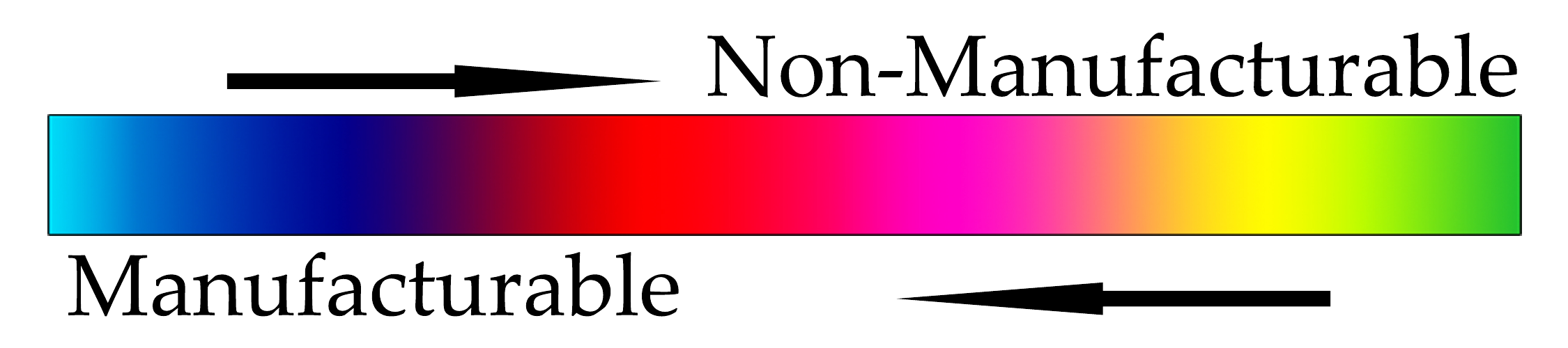}}\\
			
			\hline
		
	\end{tabular}
	
\end{table}

Using the trained DLDFM network, it is possible to obtain the localization of the feature activating the decision of the DLDFM. The 3D-GradCAM renderings for various cases are shown in the Table~\ref{Tab:GradCAM}. We have used 3D-GradCAM to visualize the results of various inputs such as manufacturable holes, non-manufacturable-holes, multiple holes in same face, and holes in multiple faces of the cube. 3D-GradCAM can localize the features that can cause the part to be non-manufacturable. For example, in Table~\ref{Tab:GradCAM}, the second example shows a CAD model with a hole, which is non-manufacturable because it is too close to one of side faces. This is a difficult example to classify based only on the information of the hole. The 3D-GradCAM rendering correctly identifies the non-manufacturable hole and as a result the DLDFM network also predicts the part to be non-manufacturable.

\section{Conclusions}
In this paper, we have developed an interpretable deep-learning-based DFM (DLDFM) framework for cyber-enabled manufacturing. The DLDFM was able to learn local features directly from the voxelized model. In addition, using the 3D-gradCAM eliminates the black box notion about CNNs; the DLDFM framework provides feedback about the source of non-manufacturability. The feedback is helpful to understand which particular local feature among various other features in a CAD geometry accounts for the non-manufacturability and possibly modify the design appropriately.
\section*{Acknowledgements}
This paper is based upon research partially supported by the National Science Foundation under Grant No. CMMI:1644441. We gratefully acknowledge the support of NVIDIA Corporation with the donation of the GTX TITAN Xp GPU used for this research. The data and code used in this paper can be found at \url{http://web.me.iastate.edu/adamlab/Index.html}.

{\small
	\bibliographystyle{IEEEtranN}
	\bibliography{CM,ML,CAD}
}
\end{document}